\documentclass[pdflatex,sn-mathphys-num]{sn-jnl}

\usepackage{graphicx}%
\usepackage{multirow}%
\usepackage{amsmath,amssymb,amsfonts}%
\usepackage{amsthm}%
\usepackage{mathrsfs}%
\usepackage[title]{appendix}%
\usepackage{xcolor}%
\usepackage{textcomp}%
\usepackage{manyfoot}%
\usepackage{booktabs}%
\usepackage{algorithm}%
\usepackage{algorithmicx}%
\usepackage{algpseudocode}%
\usepackage{listings}%
\usepackage{soul}
\usepackage{natbib}
\setcitestyle{authoryear,open={(},close={)},aysep={}}
\usepackage{arydshln}

\theoremstyle{thmstyleone}%
\theoremstyle{thmstyletwo}%

\theoremstyle{thmstylethree}%

\begin{document}

\title[Article Title]{Public Attitudes Toward ChatGPT on Twitter: Sentiments, Topics, and Occupations}


\author*[1]{\fnm{Ratanond} \sur{Koonchanok}}\email{rkoonch@iu.edu}
\equalcont{These authors contributed equally to this work.}

\author[2]{\fnm{Yanling} \sur{Pan}}\email{panyanl@iu.edu}
\equalcont{These authors contributed equally to this work.}

\author[3]{\fnm{Hyeju} \sur{Jang}}\email{hyejuj@iu.edu}

\affil*[1]{\orgdiv{Department of Human-Centered Computing}, \orgname{Indiana University–Purdue University Indianapolis}, \city{Indianapolis}, \state{IN}, \country{USA}}

\affil[2]{\orgname{Luddy School of Informatics, Computing, and Engineering}, \orgname{Indiana University–Purdue University Indianapolis}, \city{Indianapolis}, \state{IN}, \country{USA}}

\affil[3]{\orgdiv{Department of Computer Science}, \orgname{Indiana University–Purdue University Indianapolis}, \city{Indianapolis}, \state{IN}, \country{USA}}


\abstract{ChatGPT sets a new record with the fastest-growing user base, as a chatbot powered by a large language model (LLM). While it demonstrates state-of-the-art capabilities in a variety of language-generation tasks, it also raises widespread public concerns regarding its societal impact. In this paper, we investigated public attitudes towards ChatGPT by applying natural language processing techniques such as sentiment analysis and topic modeling to Twitter data from December 5, 2022 to June 10, 2023.
Our sentiment analysis result indicates that the overall sentiment was largely neutral to positive, and negative sentiments were decreasing over time. Our topic model reveals that the most popular topics discussed were \emph{Education}, \emph{Bard}, \emph{Search Engines}, \emph{OpenAI}, \emph{Marketing}, and \emph{Cybersecurity}, but the ranking varies by month. We also analyzed the occupations of Twitter users and found that those with occupations in \emph{arts and entertainment} tweeted about ChatGPT most frequently. Additionally, people tended to tweet about topics relevant to their occupation. For instance, \emph{Cybersecurity} is the most discussed topic among those with occupations related to \emph{computer and math}, and \emph{Education} is the most discussed topic among those in \emph{academic and research}. Overall, our exploratory study provides insights into the public perception of ChatGPT, which could be valuable to both the general public and developers of this technology. }




\maketitle

\section{Introduction}\label{sec1}

ChatGPT has gained widespread attention since its release and has been the subject of public discourse across various communities.
It reached an estimated 100 million monthly active users in January 2023, just two months after its release\footnote{\href{https://www.similarweb.com/blog/insights/chatgpt-25-million/}{similarweb: ChatGPT Tops 25 Million Daily Visits}}. This is a remarkable feat compared to other popular social media platforms such as TikTok, which took 9 months to reach the same milestone, Instagram, which took 2.5 years, and Pinterest, which took 3.5 years\footnote{\href{https://news.yahoo.com/chatgpt-on-track-to-surpass-100-million-users-faster-than-tiktok-or-instagram-ubs-214423357.html}{ChatGPT on track to surpass 100 million users faster than TikTok or Instagram: UBS}}.

ChatGPT is a large language model (LLM) developed by OpenAI that generates plausible texts upon prompts for a wide range of purposes such as academic writing, language translations, and coding. It has captured people’s imagination about a revolutionized future of programming \citep{chen2021evaluating, li2022competition}, writing \citep{mirowski2022co, lee2022coauthor}, and many more. ChatGPT has demonstrated its ability to boost human productivity and shift human work from the more laborious code/text generation to problem formulation and higher-level thinking.

On the other hand, ChatGPT is not without flaws \citep{borji2023categorical2}, just as its LLM predecessors. It has the propensity to generate falsehood or `hallucinate' with high confidence \citep{lin2022teaching, zhou2023navigating} such as by making up fictitious citations \citep{NatMachIntell2023AIwriting}. It is also weak at math and logic problems \citep{han2022folio}. From a societal standpoint, it becomes alarmingly disturbing considering the bias and discrimination inherent to LLMs from their training datasets \citep{jakesch2023co, lin2021truthfulqa, weidinger2021ethical, zhuo2023exploring, abid2021large}. 
Additionally, the susceptibility to misuse and abuse of the technology could facilitate the spread of disinformation and misinformation \citep{buchanan2021truth}, potentially swaying public opinion and causing real-world harm.

In addition to the impact on industries and the general public, ChatGPT has been revolutionizing academia, affecting how scientists approach research, academic writing, and critical thinking. Within one month of release, ChatGPT has been listed as an author in many academic papers \citep{o2022open, transformer2022rapamycin, Kung2022.12.19.22283643}, leading multiple journals to officially prohibit AI tools like ChatGPT from being listed as authors \citep{anon2023tools, thorp2023chatgpt}. Professional organizations, websites, and schools also took measures to restrict the usage of or block access to ChatGPT, in response to concerns about falsehood, plagiarism, and unpredictable implications for education.

Studying the social context around technology could be crucial to its future adoption and application \citep{frewer1998understanding}. Even before the ChatGPT era, studies on public opinion toward chatbots have been conducted to better comprehend the roles they play in various domains. In marketing, for example, understanding the attitude toward chatbots in mobile advertising can help companies develop and test new strategies \citep{de2021role}. Similarly, teachers' attitude toward chatbots can affect their willingness to adopt the technology as part of their work \citep{bii2018teacher}. Studies that focus on attitudes toward ChatGPT, however, are limited.  

The Twitter community is abuzz with discussions about ChatGPT, with users sharing their first-hand experiences with the technology and its potential for future use in various domains. Accordingly, we will investigate the public attitudes toward ChatGPT using tweets from both overall and domain-specific users. In particular, we ask the following research questions. 

\begin{enumerate}
    \item What is the public sentiment towards ChatGPT in tweets, and how does it change over time? 
    \item What ChatGPT-related topics do people discuss, and what are the topic-specific sentiments?
    \item What are the occupations of users who post about ChatGPT on Twitter? What topics do each occupation group discuss?

\end{enumerate}

We will first apply a sentiment analysis model to ChatGPT-related tweets to understand the overall sentiment and use a topic modeling technique to discover the most discussed topics. We will then extract occupation information from the user description associated with each tweet and analyze occupation-related patterns in ChatGPT-related topics.

\section{Related Work}

ChatGPT has been the subject of growing research since its release. Some studies examined its broad social impacts, such as its potential to be exploited for misinformation or to spread harmful stereotypes \citep{abdullah2022chatgpt}. Others focused on its potential in specific domains, such as education and healthcare \citep{dwivedi2023so, biswas2023role, tlili2023if, dempere2023impact, shoufan2023exploring}. The majority of these studies employed social science methods such as surveys and questionnaires, user experience analyses, expert opinions, and literature reviews. More relevant to our work are those that applied computational techniques to analyze large-scale social media data. We will discuss these studies in detail in the current section. 

To investigate general public discussions about ChatGPT on Twitter, the majority of studies focused on early opinions and reactions using data collected within the first few months of ChatGPT release \citep{tounsi2023exploring, korkmaz2023analyzing, haque2022think, taecharungroj2023can, leiter2023chatgpt}, ranging from 2 to 41 days. Among them, \citet{haque2022think} and \citet{taecharungroj2023can} performed topic modeling using the Latent Dirichlet Allocation (LDA) technique \citep{blei2003latent}, aiming to discover popular topics in ChatGPT-related tweets. \citet{leiter2023chatgpt} used a roBERTa-based model fine-tuned for tweet topic classification to classify tweets into pre-determined topics for analysis. 

To examine the public sentiment toward ChatGPT, \citet{tounsi2023exploring} and \citet{korkmaz2023analyzing} applied lexicon- or rule-based models to obtain sentiment scores, while \citet{haque2022think} performed per-topic sentiment analysis by manually labeling the tweets. Most similar to our study, \citet{leiter2023chatgpt} used a Transformer-based model fine-tuned for sentiment analysis to classify each tweet as positive, negative, or neutral. \citet{leiter2023chatgpt} and \citet{tounsi2023exploring} also applied emotion classification models to examine the public reactions expressed in tweets.

Studies that focused on specific areas like cybersecurity \citep{okey2023investigating}, healthcare \citep{praveen2023understanding}, or education \citep{li2023chatgpt,futterer2023chatgpt} examined tweets with relatively longer time spans, ranging from 2 to 4 months. For sentiment analysis, these studies either used the lexicon- and rule-based VADER model \citep{hutto2014vader} or a pre-trained BERT model like Twitter-roBERTa \citep{barbieri2020tweeteval}, both of which we evaluated for performance in our study. To perform topic modeling, three of these studies used BERTopic \citep{, praveen2023understanding, li2023chatgpt, futterer2023chatgpt}, while \citet{okey2023investigating} used LDA.

The studies mentioned above revealed either the overall or per-topic sentiments of the public toward ChatGPT in tweets. However, public attitudes are complex and can be difficult to understand, especially because the public is made up of individuals with different demographics. More importantly, all of the previous studies, whether they focused on a specific area or not, analyzed data that covered a relatively short time span, ranging from 2 days to 4 months. In this paper, we aim to understand public attitudes toward ChatGPT by examining Twitter data spanning 7 months since release, providing additional insights on demographic- and topic-dependent differences in public sentiments, and revealing how public opinions change as people have had more time to experiment with ChatGPT.

\begin{figure*}[!htb]
  \center
  \includegraphics[width=1\textwidth]{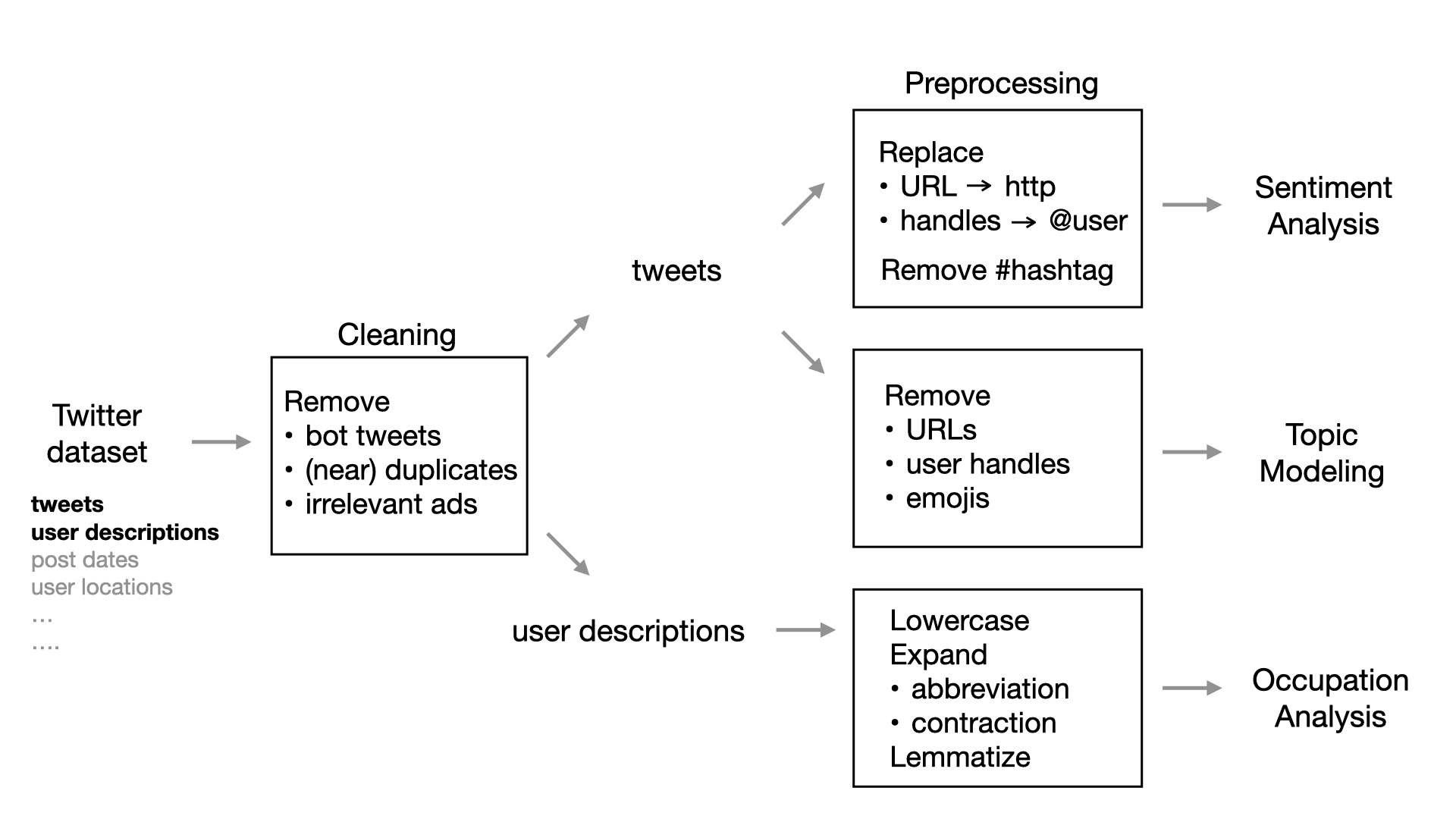}
  \caption{Overview of data cleaning and preprocessing }
  \label{data_flow}
\end{figure*}

\section{Methods}

This section describes our methodology in detail, from data acquisition to analysis methods. We first applied a set of data-cleaning methods to remove irrelevant data from the dataset. We then preprocessed the tweets by handling tweet-specific features (such as hashtags and URLs) and normalizing the language. The overview of our workflow is shown in Figure \ref{data_flow}. 

\subsection{Dataset}

We used a dataset of tweets from Kaggle that were collected using the hashtag,  \texttt{\#ChatGPT}\footnote{\url{https://www.kaggle.com/datasets/konradb/chatgpt-the-tweets/versions/172}}. This dataset has been updated daily starting from December 5, 2022, five days after the release of ChatGPT. We used version 172, which was last updated on June 10, 2023, and contained 462,980 tweets along with their corresponding user name, user description, user location, other additional hashtags, and post date. 

\subsection{Data Cleaning}
Following the Twitter data cleaning steps in \citep{loureiro2022timelms}, we removed duplicates, near-duplicates, and tweets from bots to ensure that our analysis of tweets would be more accurate and informative.  Bots were defined as the 1\% most active users who posted most frequently and were also more likely to post repeated content. Near-duplicates were identified using MinHash \citep{broder1997resemblance}, a technique for estimating similarity between documents. Additionally, we used keyword filtering to remove irrelevant tweets, such as crypto and sales ads, which contained popular hashtags like \#ChatGPT only for marketing purposes. The resulting dataset of 314,645 tweets was used for our analysis.

\subsection{Sentiment Analysis} \label{sentiment_method}
To answer the first research question regarding the general sentiment towards ChatGPT, we first preprocessed the texts.  Following the steps in \citep{loureiro2022timelms}, we replaced user handles and URL links with generic placeholders, namely \emph{@user} and \emph{http}. Note that we neither lowercased the text nor removed emojis for this task, because they could provide sentiment information and were included for the training of sentiment analysis models we would use.

We selected three sentiment analysis models that are specifically suitable for Twitter data: VADER \citep{hutto2014vader}, Twitter-roBERTa \citep{barbieri2020tweeteval}, and XLM-T \citep{barbieri2022xlm}. VADER is a lexicon- and rule-based model, while Twitter-roBERTa and XLM-T are Transformer-based models trained on tweets and fine-tuned for sentiment analysis. XLM-T is different from Twitter-roBERTa mainly in its multilingual capacity because it was trained on tweets in eight languages, while Twitter-roBERTa was trained on English tweets only\footnote{XML-T achieved better performance on sentiment analysis of English tweets when trained on multilingual tweets than when trained solely on English tweets.}.
All three models are publicly available as a Python package or on Hugging Face. For our analysis, we did not re-train or fine-tune the models. A more detailed description of the three models is included in Appendix \ref{appendix:models}.

We finalized the best model for our study by evaluating the performance of these models against the ground truth in a subset of our dataset. To generate the ground truth, we randomly selected 1,000 tweets from the cleaned dataset and manually labeled their sentiments. As shown in Appendices \ref{appendix:model_eval} and \ref{appendix:sentient_distribution_1000}, the XLM-T model performed the best among the three and thus was used for the sentiment analysis on our full dataset. 


\subsection{Topic Modeling}
\label{sec:topic-modeling}

To answer the second research question, we further preprocessed the tweets in section 3.3 by removing emojis, as well as the generic placeholders of URLs and user handles (\emph{http} and \emph{@user}), because they were not necessary for the subsequent steps. 

We extracted topics from tweets using BERTopic \citep{grootendorst2022bertopic}, a topic modeling technique that leverages the clustering of Transformer-based document embeddings. The parameters for BERTopic were set as `all-mpnet-base-v2' for \emph{embedding\_model}, `(1,2)' for \emph{ngram\_range}, and `english' for \emph{stop\_words}. Eventually, a topic was assigned to each tweet. We labeled each topic based on the most frequent term in the corresponding topic representation. To explore the most frequent topics for each month, we grouped tweets based on their post dates and ranked the topic frequency within each month. 

To probe deeper into our analysis, we chose a set of potentially controversial topics, aiming to explore both the sentiment towards each topic and its change over time. Building upon the sentiment analysis conducted in Section \ref{sentiment_method}, we delved deeper into these chosen topics. Our sentiment analysis model, which assigned a unique sentiment label to each tweet, allowed us to dissect the sentiment distribution specific to each topic.

To examine sentiment changes over time, we grouped tweets based on their post date. The first group encompassed tweets from weeks 1 to 12, while the second group covered weeks 13 to 25. Leveraging the sentiment labels from Section \ref{sentiment_method}, we calculated the sentiment distribution for each time frame and subsequently tracked the changes between them.

\subsection{Occupation Extraction and Analysis}
\label{occupation method} 

To answer the final research question, we grouped tweets based on user occupations by combining information from the user description associated with each tweet and three curated lists of occupation-related terms primarily based on the Standard Occupational Classification (SOC) of the U.S. Bureau of Labor Statistics\footnote{\url{https://www.bls.gov/soc/2018/major_groups.htm}}. We will describe how we use the SOC in this section and leave a detailed description of other supplementary lists in Appendix \ref{appendix:occupation}.

First, the SOC contains 23 occupation groups with each group encompassing multiple occupation titles. For example, the \emph{Computer and Mathematical Occupations} group includes occupation titles such as software developer, data scientist, and mathematician. We expanded the list to be more inclusive by incorporating abbreviations and occupations that frequently appear on Twitter but are not initially included in the SOC. For example, we added abbreviations such as CEO, prof, and tech as well as occupations such as student, founder, and YouTuber. We further simplified the occupation groups by merging the 13 groups with frequency less than 1\% in our dataset into a single group called \emph{others}. We also merged \emph{Healthcare Practitioners and Technical Occupations} and \emph{Healthcare Support Occupations} into a single group called \emph{healthcare}. 

Next, we preprocessed the text in user descriptions of each tweet and obtained the corresponding unigrams and bigrams. User descriptions generally contained users' self-described information, frequently including their occupations. We preprocessed the descriptions by removing URLs, user handles, emojis, stop words, digits, and punctuations, as well as expanding abbreviations, expanding contractions, applying lowercase, and applying lemmatization. The python packages, \texttt{nltk}, \texttt{contractions}, and \texttt{spaCy} were used for processing stop words, contractions, and lemmatization, respectively. 

Finally, we extracted a unique occupation for each tweet by matching the unigrams and bigrams of user descriptions to the curated lists of occupations. Detailed descriptions of the matching steps are included in Appendix \ref{appendix:occupation}. 

Leveraging the identified occupations, we categorized the entire tweet corpus by occupation. Subsequently, we employed the topics discovered from Section \ref{sec:topic-modeling} to uncover the most prominent topics within each occupation group.

\section{Results}\label{sec2}

We present the results of our sentiment analysis, topic modeling, and occupational analysis. We analyze both the overall trend and changes over different time periods.

\subsection{Sentiment Analysis}

Our sentiment analysis of 314,645 tweets reveals a generally positive public perception of ChatGPT. As shown in Figure 1, the breakdown is: 35.28\% positive, 45.79\% neutral, and 18.92\% negative.

\begin{figure}[!hbt]
  \center
  \includegraphics[width=.5\textwidth]{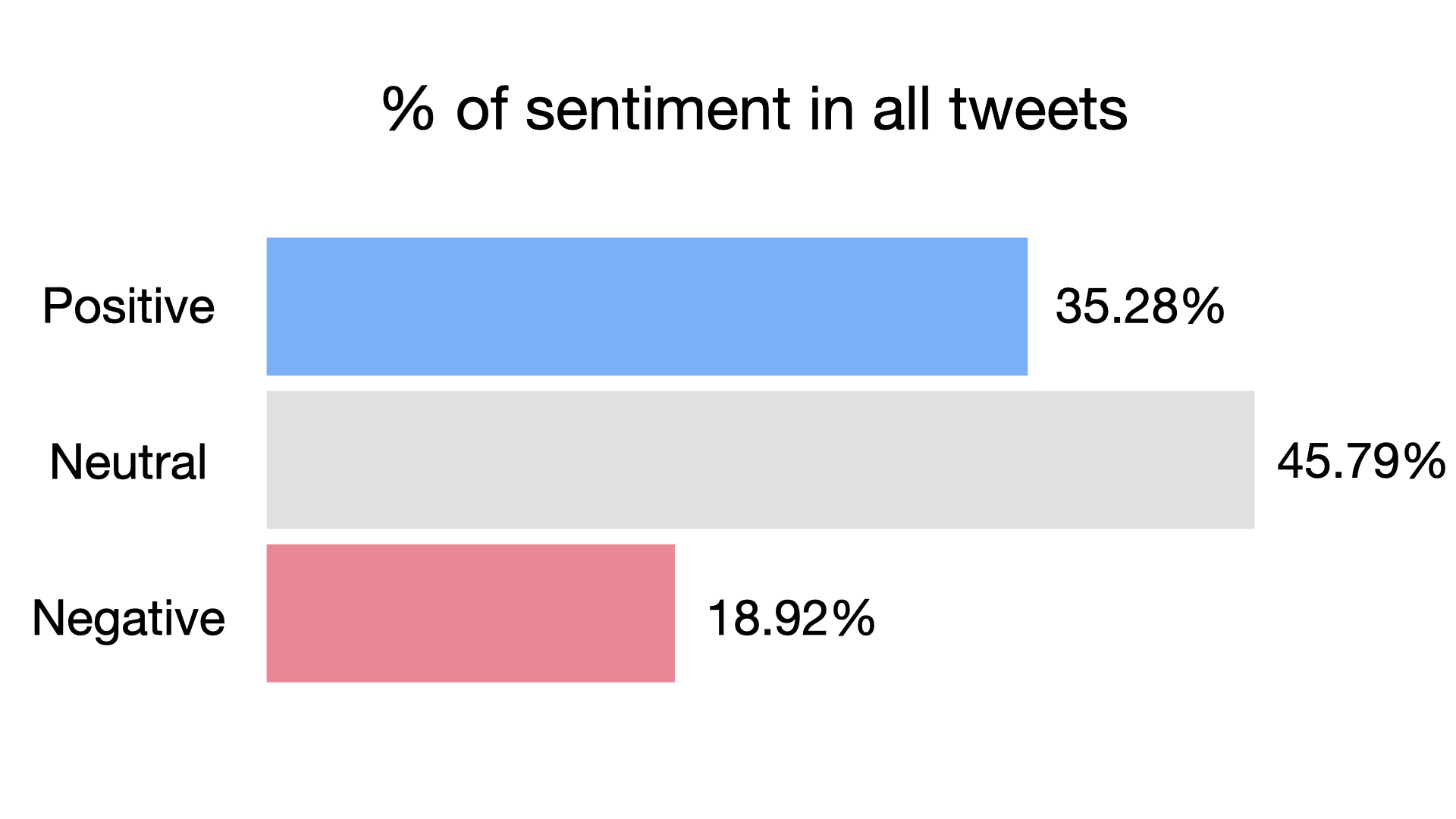}
  \caption{Distribution of positive, neutral, and negative sentiments of the full dataset predicted by XLM-T }
  \label{sentiment_analysis_full}
\end{figure}

\begin{figure}[!hbt]
  \center
  \includegraphics[width=.5\textwidth]{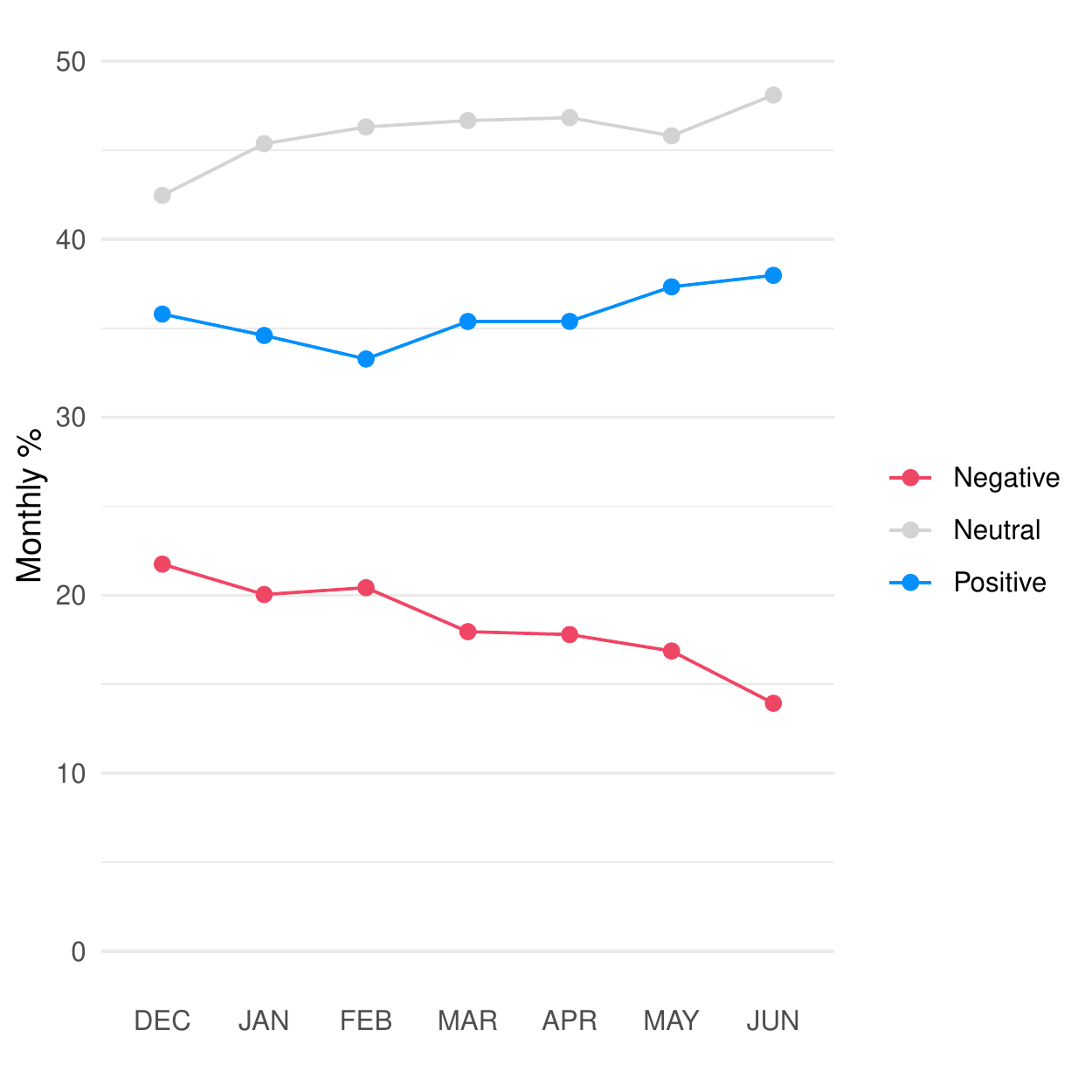}
  \caption{ Sentiment distribution by month from December 2022 to June 2023}
  \label{sentiment_analysis_byMonth}
\end{figure}

Figure \ref{sentiment_analysis_byMonth} illustrates the changes in sentiment with respect to the tweets from December 2022 to Jun 2023. Throughout the months, positive tweets increased from 35.79\% in December to 37.97\% in June. Similarly, neutral tweets increased from 42.46\% to 48.09\%. In contrast, negative tweets decreased from 21.75\% to 13.94\%.

\subsection{Topic Modeling Analysis} \label{sec:topic-modeling result}
\subsubsection{Most frequent topics}
\label{Most Frequent Topics}

\begin{table*}[!hbt]
    \centering
    \begin{tabular}{lcl}
    \hline
    Topic     & Count & Example\\
    \hline

Education & 7,994 & Teachers are using ChatGPT more than students...  \\
Bard  & 7,207 & Google Bard vs ChatGPT: Which Is the Best AI Chatbot? \\
Search Engines    & 6,817 & Will ChatGPT Replace Google Search Engine? \\
OpenAI   & 6,105 & OpenAI shares their GPT best practices!\\
Marketing & 5,726 & How are you going to use ChatGPT in content marketing?\\
Cybersecurity & 5,139 & ChatGPT is a threat to privacy.\\

    \hline
    \end{tabular}
    \caption{Most frequent topics with tweet counts and examples }
    \label{tab:topic_freq}
\end{table*}

\begin{table*}[!hbt]
    \centering
    \begin{tabular}{llllllll}

\hline
Dec 2022          & Jan 2023        & Feb 2023   & Mar 2023        & Apr 2023       & May 2023        & Jun 2023 \\
\hline
Search          & Education     & Search    & OpenAI        & Italy         & Bard          & Bard \\
OpenAI          & Search        & Bard      & Education     & Cybersecurity & OpenAI        & LLMs \\
Education       & OpenAI        & Bing      & Marketing     & Education     & Education     & Marketing \\
Writing         & Marketing     & Education & Cybersecurity & Marketing     & Marketing     & Education \\
Twitter         & Writing       & Marketing & GPT3          & Healthcare    & Cybersecurity & Cybersecurity\\

    \hline
    \end{tabular}
    \caption{Most frequent topics for each month between Dec 2022 and June 2023 }
    \label{tab:topic_freq_month}
\end{table*}

Our BERTopic model identified 101 topics, along with a significant group of outliers (48.89\% of tweets). While merging outliers with other topics was possible, we chose not to do so in order to preserve the cluster quality. 

Among the 101 identified topics, the most frequent was \emph{AI and ChatGPT} with 22,836 tweets (7.26\%). We excluded this topic from the subsequent analysis to focus on tweets discussing specific applications and aspects of ChatGPT, rather than ChatGPT itself. An example tweet from this topic is ``ChatGPT knows it best!'' While tweets like this can be valuable for sentiment analysis, they offer limited insights into the diverse range of topics surrounding ChatGPT. 

Excluding this topic about \emph{AI and ChatGPT}, the remaining 100 topics provide a rich landscape of public interest. The most frequent topics include \emph{Bard}, \emph{Search Engines}, \emph{OpenAI}, \emph{Marketing}, and \emph{Cybersecurity}. Table \ref{tab:topic_freq} shows the top 6 topics by tweet count, along with representative examples for each.

Table \ref{tab:topic_freq_month} further explores the evolution of these topics, displaying the most frequent topics on a monthly basis from December 2022 to June 2023. This allows us to observe which topics gained or lost traction over time, offering valuable insights into the shifting dynamics of public perception.

\subsubsection{Sentiments for selected topics}

\begin{figure*}[!htb]
  \center
  \includegraphics[width=1\textwidth]{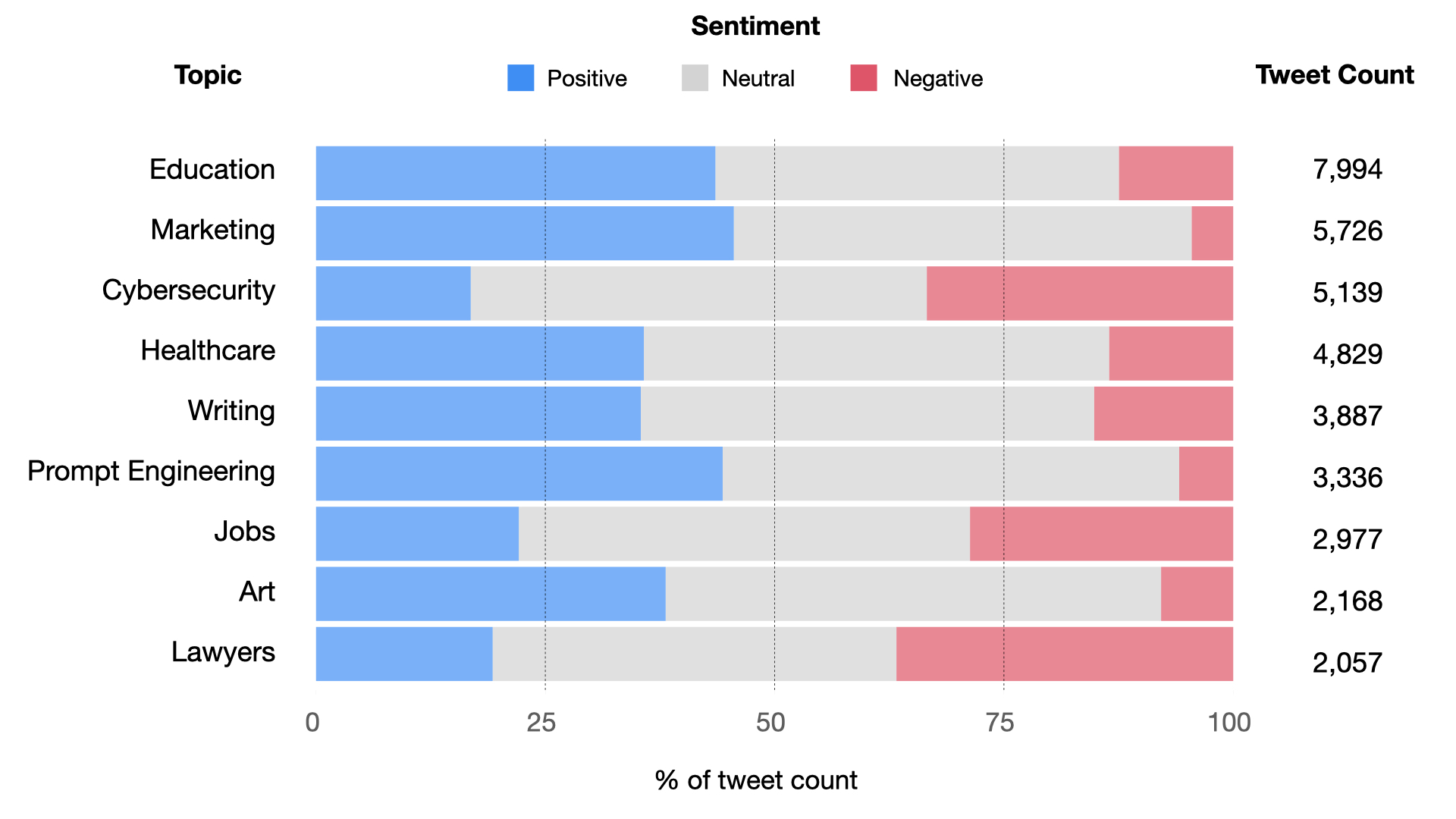}
  \caption{Percentage distribution of negative, neutral, and positive sentiments for selected topics}
  \label{fig:sentiment_dist_selectedTopics}
\end{figure*}

Figure \ref{fig:sentiment_dist_selectedTopics} reveals the sentiment distribution for each of the potentially controversial topics selected in our analysis. The tweets about certain topics such as \emph{Education}, \emph{Marketing}, \emph{Healthcare}, \emph{Writing}, \emph{Prompt Engineering}, and \emph{Art} are more positive whereas the tweets about \emph{Cybersecurity}, \emph{Jobs}, and \emph{Lawyers} are more negative.

\begin{figure*}[!htb]
  \center
  \includegraphics[width=1\textwidth]{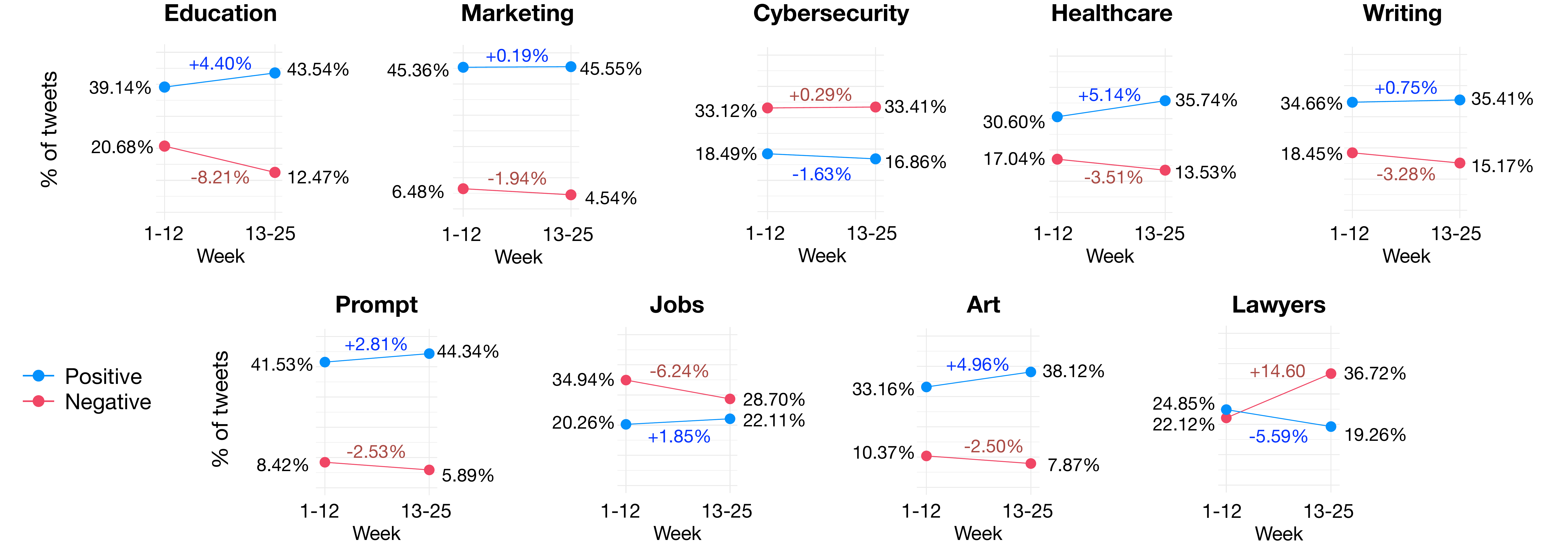}
  \caption{Changes in sentiment for selected topics after 12 weeks} 
  \label{fig:selectedTopics_12week}
\end{figure*}

In addition, we track the percentage shifts in positive and negative tweets for each topic. These in-depth trends are shown in Figure \ref{fig:selectedTopics_12week}.

\begin{table}[!hbt]
    \centering
    \begin{tabular}{lrr}
    \hline
    Occupation    & Count & \% Total\\
    \hline

arts and entertainment    &  42,108 & 13.38\\
management                &  28,861 &  9.17\\
computer and math         &  27,763 &  8.82\\
business and finance      &  27,619 &  8.78\\
engineering               &  16,770 &  5.33\\
academic and research     &  16,497 &  5.24\\
education                 &   9,190 &  2.92\\
sales                      &   7,277 &  2.31\\
healthcare                &   4,651 &  1.48\\
student                   &   3,444 &  1.09\\
\hdashline
others                    &   5,797 &  1.84\\
not available             & 124,668 & 39.62\\

    \hline
    \end{tabular}
    \caption{Occupation distribution. \emph{not available} includes tweets that did not contain a user description and that failed in occupation extraction. \emph{others} encompasses occupations with frequency less than 1.00\%.}
    \label{tab:occupation_dist}
\end{table}

\subsection{Occupation Analysis}
\subsubsection{Identified Occupations}
We successfully identified occupation information from 60.38\% (189,977 tweets) of the dataset. For the remaining 39.62\% (124,668 tweets), either user descriptions were absent (5.83\% or 18,338 tweets) or our extraction method did not find matching occupation titles in the descriptions (33.79\%  or 106,330 tweets). Table \ref{tab:occupation_dist} presents the frequency of identified occupations based on unique tweets.

\subsubsection{Topic Trends across Occupations}

We delved into the most popular topics within different occupations, showing how they might differ from the most popular topics in the overall tweets observed in Table \ref{tab:topic_freq}. To achieve this, we analyzed the topics identified by the BERTopic in Section \ref{sec:topic-modeling result}, focusing specifically on those found within tweets from users with each defined occupation. We excluded the ubiquitous \emph{AI and ChatGPT} topic to maintain consistency with our previous findings. The key findings are summarized in Table \ref{tab:topic_freq_occ}.

\begin{table*}[!hbt]
    \centering
    \begin{tabular}{r|lllll}

\hline
Occupation & \multicolumn{5}{c}{Top 5 topics}\\
\hline
arts and entertainment  & Writing       & Bard          & Education     & Search        & Marketing\\
Management              & Marketing     & Search        & Bard          & Education     & OpenAI\\
computer and math       & Cybersecurity & OpenAI        & Bard          & Search        & LLMs\\
business and finance    & Crypto        & Marketing     & Bard          & Search        & OpenAI     \\
engineering             & Bard          & Search        & OpenAI        & Bing          & Cybersecurity     \\
academic and research   & Education     & Healthcare    & References    & LLMs   & Search\\
education               & Education     & Healthcare    & Writing       & Search        & Bard \\
sales                    & Marketing     & Search        & SEO           & Bard          & OpenAI \\
healthcare              & Healthcare    & Education     & Bard          & References    & Poem \\
student                 & Education     & Bard          & Search        & LLMs   & Healthcare \\

    \hline
    \end{tabular}
    \caption{Most popular topics in each occupation ordered by frequency from left to right}
    \label{tab:topic_freq_occ}
\end{table*}

\section{Discussion}

To study the public attitudes toward ChatGPT, we applied a Transformer-based sentiment analysis model to extract sentiment from tweets and topic modeling techniques to group tweets based on topics. We also grouped tweets based on occupation using the user description in conjunction with a list of official occupations.

\textbf{Overall, the tweet sentiment is mildly positive}, with 35.28\% of tweets being positive, 45.79\% being neutral, and 8.92\% being negative. This result is consistent with other works, albeit with minor differences. For instance, our result shows a lower percentage of neutral tweets compared to the work by \citet{leiter2023chatgpt}, which reports 29.9\% positive, 52.2\% neutral, and 17.9\% negative tweets. This difference could be due to our stricter standard for removing bot tweets and near duplicates, resulting in a different sentiment distribution.

The most frequently discussed ChatGPT-related topics were \textbf{\emph{Education}, \emph{Bard}, \emph{Search Engines}, \emph{OpenAI}, \emph{Marketing}, and \emph{Cybersecurity}}. Tweets about these topics generally discussed how ChatGPT could initiate technical and societal disruption. Conversations about \emph{Education} frequently revolved around whether ChatGPT could be used as part of the learning experience and job environment. For \emph{Bard} and \emph{Search Engines}, the discussions centered around how ChatGPT could replace or be integrated as part of search engines.

The results of analyzing the most frequent topics by month indicate that \textbf{people not only discussed applications of ChatGPT but also talked about ChatGPT-related events}. For instance, in February, \emph{Bard} and \emph{Bing} entered the top-5 list for the first time. This was the month when Google launched its own generative AI named Bard, while Microsoft announced the integration of ChatGPT into its search engine Bing. Similarly, in March, people started mentioning GPT-3 more frequently after the release of GPT-4. In April, the reversion of the ban on ChatGPT in Italy corresponded to Italy being the most frequent topic for the month.

Our analysis of tweet sentiment revealed \textbf{positive trends toward ChatGPT in specific topics: \emph{Education}, \emph{Marketing}, \emph{Healthcare}, \emph{Writing}, \emph{Prompt Engineering} and \emph{Art}}. These topics show the domains that hold promise for utilizing ChatGPT's capabilities. In education, for instance, the growing integration of technology has sparked discussions about ChatGPT's potential to enhance learning. Similarly, the proven effectiveness of language processing in marketing since the chatbot era (e.g., \citep{cheng2022customer}) motivates professionals to explore ChatGPT's lucrative applications. The high volume of positive tweets endorsing ChatGPT's marketing usefulness reflects this optimism. Furthermore, \emph{Prompt Engineering}, the art of crafting optimal inputs for ChatGPT, garnered mainly positive sentiment, likely due to its increasing importance across diverse professions.

While the sentiment towards ChatGPT remained positive across \emph{Education}, \emph{Healthcare}, \emph{Prompt Engineering}, \emph{Art}, \emph{Marketing}, and \emph{Writing}, \textbf{its temporal evolution varied}. For \emph{Education}, \emph{Healthcare}, \emph{Prompt Engineering}, and \emph{Art}, the percentage of positive tweets increased while that of negative tweets decreased over time. For \emph{Marketing} and \emph{Writing}, the positive tweets stayed nearly constant while negative tweets decreased.

On the other hand, \textbf{\emph{Cybersecurity} and \emph{Jobs} were the only topics with predominantly negative sentiment in the initial phase} (the first 12 weeks). This negativity might stem from anxieties about disruption, as various applications of ChatGPT could alter the job market and raise cybersecurity risks \citep{frith2023chatgpt,chow2023impact}. However, attitudes might shift as people understand ChatGPT's actual capabilities better. Notably, these topics had minimal sentiment changes between periods.

\textbf{The topic \emph{Lawyers} uniquely experienced both an increase in negative and a decrease in positive tweets}. This likely reflects hesitation about legal tasks entrusted to ChatGPT. For example, \citet{choi2023chatgpt} showed its performance falls short of law school graduates, potentially fueling distrust.

Interestingly, our findings on the sentiment towards ChatGPT in \emph{Education}, \emph{Healthcare}, and \emph{Cybersecurity} largely mirror those of previous studies \citep{tubishat2023sentiment, praveen2023understanding, okey2023investigating}. Positive trends emerge in \emph{Education} and \emph{Healthcare}, while concerns dominate the discussion in \emph{Cybersecurity}. Note that this comparison is limited to these three topics due to the limited availability of sentiment analysis results in existing literature.


The occupation analysis reveals that the \emph{arts and entertainment} group, comprising writers, journalists, musicians, and dancers, posts more frequently about ChatGPT. In addition, grouping topics by occupation shows \textbf{a tendency for individuals to discuss their profession-related topics}, which aligns with our initial expectations. For example, as shown in Table \ref{tab:topic_freq_occ}, the topic \emph{Education} was the most frequent topic for users whose jobs are in the groups of \emph{academic and research}, \emph{education}, and \emph{student}.

\section{Limitation}

Some limitations should be taken into consideration when interpreting our results. First, our analysis is limited to tweets collected before June 2023, due to changes in Twitter's data access policy. This may not fully capture the latest trends and discussions. Second, our occupation extraction relied on user-provided descriptions, potentially excluding individuals without descriptions or with professions not included in our list. The list itself, while extensive, may not encompass all possible occupations. Finally, we assumed one unique occupation per tweet, which might not reflect the full complexities of individuals' career identities. This simplifies the analysis but requires caution when interpreting results.

\section{Conclusion}\label{sec13}

Our study investigated public perceptions of ChatGPT, offering valuable insights for both the general users interested in ChatGPT and the developers of ChatGPT-related technology. Users can glean a broader understanding of the technology, informing their usage choices. Developers can gain crucial context, enabling them to tailor future iterations. Moreover, by revealing how demographics influence attitudes, our findings hold the potential for optimizing design, marketing, and responsible development of this technology, ultimately contributing to a nuanced understanding of its societal impact.

\backmatter



\section*{Declarations}

\bmhead{Funding}
There is no funding directly associated with this study to declare.

\bmhead{Competing interests}
The authors have no competing interests as defined by Springer, or other interests that might be perceived to influence the results and/or discussion reported in this paper.

\bmhead{Ethics approval and consent to participate} 
Not applicable
\bmhead{Consent for publication }
Not applicable
\bmhead{Data availability }
Our dataset was retrieved from \url{https://www.kaggle.com/datasets/konradb/chatgpt-the-tweets/versions/172}.

\bmhead{Materials availability}
Not applicable
\bmhead{Code availability}
Our code is available at \url{https://github.com/yanlingpan/chatgpt_sentiment.git}.

\bmhead{Author contribution}
All authors contributed to the conception and design of the study. R.K. and Y.P. performed data collection, preprocessing,
and all analyses. R.K. and Y.P. wrote the first draft of the manuscript. All authors contributed to the manuscript revision, read, and approved the submitted version.


\begin{appendices}

\section{Sentiment Analysis Models}
\label{sec:appendix}

\subsection{Model Choice}
\label{appendix:models}
\textbf{VADER} (Valence Aware Dictionary for sEntiment Reasoning) \citep{hutto2014vader} is a lexicon- and rule-based model for sentiment analysis. It combines a list of gold standard lexicons that are tuned to sentiments in tweets with the grammatical and syntactical rules in human language that capture expression and emphasis of sentiment intensity. The model also includes special handling of negation, emojis, and emoticons (e.g.:D). This model was selected for comparison because it represents the state-of-the-art rule-based on sentiment analysis of tweets. VADER returns a sentiment score in the range $[-1, 1]$. We classified sentiments using thresholds recommended by the authors \footnote{\url{https://github.com/cjhutto/vaderSentiment}}:

\begin{center}
    \begin{tabular}{l l}
        Negative & $<=-0.05$ \\
        Neutral &  $(-0.05,0.05)$ \\
        Positive & $>=0.05$ 
    \end{tabular}
\end{center}

\textbf{Twitter-roBERTa} \citep{barbieri2020tweeteval} is a Transformer-based model trained on 124M tweets from January 2018 to December 2021, and finetuned for sentiment analysis with the TweetEval benchmark \citep{barbieri2020tweeteval}. It directly classifies input as positive, negative, or neutral.

\textbf{XLM-T} \citep{barbieri2022xlm} is a Transformer-based model very similar to Twitter-roBERTa, except that it has multilingual capacity, because it is trained on 198M tweets in eight languages (Ar, En, Fr, De, Hi, It, Sp, Pt). Therefore, XLM-T may have an advantage over the other two models for our dataset, in which some tweets contain non-English text that we did not remove.


\subsection{Model Evaluation}
\label{appendix:model_eval}

To obtain the best possible performance from each model, we tested additional text preprocessing steps: 1) remove hashtag symbol (\#), 2) remove hashtag sequence (sequence of more than two hashtags that are not part of a sentence), 3) both 1) and 2). For VADER, we also compared obtaining the sentiment score of the entire input with averaging sentiment scores of individual sentences of the input.

The macro-averaged recall is used to evaluate model performance, following the convention for the SemEval subtask of sentiment analysis in Twitter \citep{rosenthal2019semeval}.
\label{appendix:model_evaluation}
Table \ref{tab:sample_result} shows model performance on 1,000 manually labeled samples from the cleaned dataset. 

Averaging scores from individual sentences slightly improved VADER performance, but neither removing the hashtag sequence nor removing the hashtag symbol had a further impact. This makes sense, because the hashtag words are likely not on the list of sentiment lexicon in the model, and thus their inclusion or exclusion exerts no effect on the sentiment score. For Twitter-roBERTa, both removing the hashtag sequence and removing the hashtag symbol improved model performance, respectively. Applying both achieved the best performance with a recall score of 65.9. XLM-T without additional preprocessing achieves a higher score than the best performance of Twitter-roBERTa. Similar to Twitter-roBERTa, both removing the hashtag sequence and removing the hashtag symbol improved XLM-T performance, but removing both achieves the same best score of 67.6 as removing the hashtag symbol.

Based on these results, we will use XLM-T for sentiment analysis on the full dataset, and remove both hashtag sequence and hashtag symbol in the preprocessing step.

\begin{table}[!hbt]
    \centering
    \begin{tabular}{lcc}
    \hline
        Classifier \& Preprocessing
        &Recall\\
    \hline
    \textbf{Random}  & 32.8 \\
    \hline
    \textbf{VADER}  \\ 
             none & 50.4 \\
             avg. score & 50.9 \\ 
            avg. score +  remove sequence & 51.0 \\ 
            avg. score +  remove symbol & 50.9 \\
    \hline
    \textbf{Twitter-roBERTa} \\
            none & 62.4 \\
            remove sequence & 64.5 \\ 
            remove symbol & 64.0 \\ 
            remove sequence  + remove symbol & 65.9 \\
    \hline
    \textbf{XLM-T} \\
            none & 67.1 \\
            remove sequence & 67.4 \\ 
            remove symbol & \textbf{67.6} \\ 
            remove sequence  + remove symbol & \textbf{67.6} \\
    \hline
    \end{tabular}
    \caption{Performance of sentiment analysis models on 1,000 manually labeled random samples from cleaned
dataset using different text preprocessing.}
    \label{tab:sample_result}
\end{table}

\subsection{Sentiment Distribution of 1,000 Samples}
\label{appendix:sentient_distribution_1000}

\begin{figure}[!htb]
  \center
  \includegraphics[width=.5\textwidth]{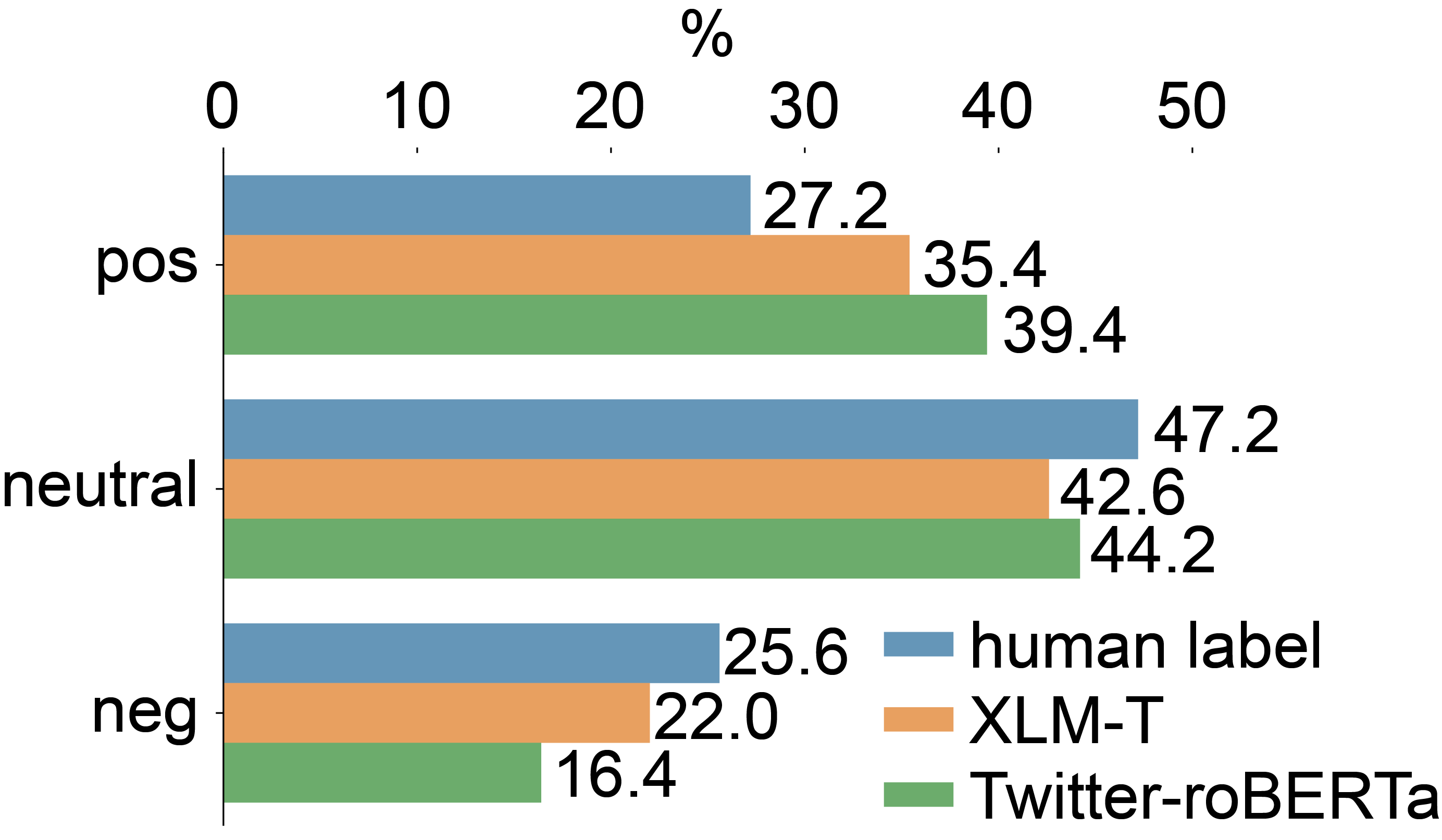}
  \caption{Distribution of positive (pos), neutral, and negative (neg) sentiments of 1,000 samples by human label and prediction by XLM-T and Twitter-roBERTa.}
\label{fig:sentient_distribution_1000sample2}
\end{figure}

To further understand the performance of Twitter-roBERTa and XLM-T, we compared the sentiment distribution of the 1,000 samples labeled by humans and those predicted by the models. As shown in Figure \ref{fig:sentient_distribution_1000sample2}, both the distributions of sentiments predicted by Twitter-roBERTa and XLM-T are different from those labeled by humans, and the distributions are also different between Twitter-roBERTa and XLM-T. Twitter-roBERTa predicts the highest percentage of positive sentiment, the lowest percentage of negative sentiment, and a similar percentage of neutral. The distribution of XLM-T is more similar to the human-labeled sentiment distribution, compared to that of Twitter-roBERTa, suggesting that using XLM-T on our dataset may achieve a more accurate result, in addition to its better evaluation performance shown in \ref{appendix:model_evaluation}. However, the difference in sentiment distribution between XLM-T and human labels also indicates that applying XLM-T to our dataset will likely underestimate the proportion of negative sentiment while overestimating that of positive sentiment.

\section{Curated Lists for Occupation Matching}
\label{appendix:occupation}

As briefly described in \ref{occupation method}, we performed occupation extraction by matching unigrams and bigrams in the user description associated with each tweet to three curated lists of occupation-related terms: \textbf{title-to-occupation}, \textbf{modifier-to-occupation}, and \textbf{keyword-to-occupation}. Here we describe in more detail how the lists were curated and how they were used during matching to obtain a unique occupation for each tweet.

The \textbf{title-to-occupation} list was curated from the SOC as described in \ref{occupation method}. It contained 329 occupation titles in 11 occupation groups and can be used to map an occupation title (e.g. software developer) to an occupation (e.g. \emph{Computer and Math}). We observed that among the 329 occupation titles, 293 of them could be unambiguously mapped to a unique occupation group using an unigram of the title, while 16 unigram titles were ambiguous and could be mapped to multiple occupation groups. For example, the ambiguous unigram title \emph{engineer} could be mapped to either \emph{computer and math} or \emph{engineering}. We also observed that the modifier words (e.g. software vs. civil) of these ambiguous titles were sufficient to disambiguate them among occupation groups. Therefore, we collected a \textbf{modifier-to-occupation} list of such occupation modifier words to help disambiguate titles that could belong to multiple occupations.

In addition, considering the nature of user description on Twitter, people do not always include occupation titles as they appear on the SOC; instead, they describe what they do using either variations of action words such as trading, editing, and coding, or modifier words such as biomedical, legal, and cybersecurity. Indeed, using the \textbf{title-to-occupation} and \textbf{modifier-to-occupation} lists, we extracted occupation from 48.35\% of our dataset, but we were able to extract an additional 12.02\% if we also matched for occupation-related keywords. Therefore, we curated a \textbf{keyword-to-occupation} list for cases where occupation extraction failed using only the \textbf{title-to-occupation} and \textbf{modifier-to-occupation} lists.

We then obtained an ordered list of unigrams and bigrams in the original order they appeared in the user description. This is because we considered the self-described information more important if it appeared earlier and should be prioritized during matching. Also, we included unigrams and bigrams, but not trigrams, etc., because based on our analysis, bigrams were sufficient to distinguish all occupation groups in our curated lists, and including trigrams did not result in more accurate extraction or higher extraction rate.

During matching, we first matched the ordered list of unigrams and bigrams to the \textbf{title-to-occupation} list and disambiguated using the \textbf{modifier-to-occupation} list when necessary. Once we successfully matched a title in \textbf{title-to-occupation}, the matched occupation was assigned to the associated tweet, and the rest of the ordered list was not examined. In other words, we assigned an occupation at the first match of a title. If no title was matched at the end of the ordered list, we performed an extra step to match the same ordered list of unigrams and bigrams to the \textbf{keyword-to-occupation} list, but instead of assigning an occupation at the first match, we obtained a unique occupation by casting a majority vote from the list of matched occupations to achieve higher accuracy of occupation extraction.

\end{appendices}


\bibliography{ref}


\begin{thebibliography}{51}
\ifx \bisbn   \undefined \def \bisbn  #1{ISBN #1}\fi
\ifx \binits  \undefined \def \binits#1{#1}\fi
\ifx \bauthor  \undefined \def \bauthor#1{#1}\fi
\ifx \batitle  \undefined \def \batitle#1{#1}\fi
\ifx \bjtitle  \undefined \def \bjtitle#1{#1}\fi
\ifx \bvolume  \undefined \def \bvolume#1{\textbf{#1}}\fi
\ifx \byear  \undefined \def \byear#1{#1}\fi
\ifx \bissue  \undefined \def \bissue#1{#1}\fi
\ifx \bfpage  \undefined \def \bfpage#1{#1}\fi
\ifx \blpage  \undefined \def \blpage #1{#1}\fi
\ifx \burl  \undefined \def \burl#1{\textsf{#1}}\fi
\ifx \doiurl  \undefined \def \doiurl#1{\url{https://doi.org/#1}}\fi
\ifx \betal  \undefined \def \betal{\textit{et al.}}\fi
\ifx \binstitute  \undefined \def \binstitute#1{#1}\fi
\ifx \binstitutionaled  \undefined \def \binstitutionaled#1{#1}\fi
\ifx \bctitle  \undefined \def \bctitle#1{#1}\fi
\ifx \beditor  \undefined \def \beditor#1{#1}\fi
\ifx \bpublisher  \undefined \def \bpublisher#1{#1}\fi
\ifx \bbtitle  \undefined \def \bbtitle#1{#1}\fi
\ifx \bedition  \undefined \def \bedition#1{#1}\fi
\ifx \bseriesno  \undefined \def \bseriesno#1{#1}\fi
\ifx \blocation  \undefined \def \blocation#1{#1}\fi
\ifx \bsertitle  \undefined \def \bsertitle#1{#1}\fi
\ifx \bsnm \undefined \def \bsnm#1{#1}\fi
\ifx \bsuffix \undefined \def \bsuffix#1{#1}\fi
\ifx \bparticle \undefined \def \bparticle#1{#1}\fi
\ifx \barticle \undefined \def \barticle#1{#1}\fi
\bibcommenthead
\ifx \bconfdate \undefined \def \bconfdate #1{#1}\fi
\ifx \botherref \undefined \def \botherref #1{#1}\fi
\ifx \url \undefined \def \url#1{\textsf{#1}}\fi
\ifx \bchapter \undefined \def \bchapter#1{#1}\fi
\ifx \bbook \undefined \def \bbook#1{#1}\fi
\ifx \bcomment \undefined \def \bcomment#1{#1}\fi
\ifx \oauthor \undefined \def \oauthor#1{#1}\fi
\ifx \citeauthoryear \undefined \def \citeauthoryear#1{#1}\fi
\ifx \endbibitem  \undefined \def \endbibitem {}\fi
\ifx \bconflocation  \undefined \def \bconflocation#1{#1}\fi
\ifx \arxivurl  \undefined \def \arxivurl#1{\textsf{#1}}\fi
\csname PreBibitemsHook\endcsname

\bibitem[\protect\citeauthoryear{Chen et~al.}{2021}]{chen2021evaluating}
\begin{botherref}
\oauthor{\bsnm{Chen}, \binits{M.}},
\oauthor{\bsnm{Tworek}, \binits{J.}},
\oauthor{\bsnm{Jun}, \binits{H.}},
\oauthor{\bsnm{Yuan}, \binits{Q.}},
\oauthor{\bsnm{Pinto}, \binits{H.P.d.O.}},
\oauthor{\bsnm{Kaplan}, \binits{J.}},
\oauthor{\bsnm{Edwards}, \binits{H.}},
\oauthor{\bsnm{Burda}, \binits{Y.}},
\oauthor{\bsnm{Joseph}, \binits{N.}},
\oauthor{\bsnm{Brockman}, \binits{G.}}, et al.:
Evaluating large language models trained on code.
arXiv preprint arXiv:2107.03374
(2021)
\end{botherref}
\endbibitem

\bibitem[\protect\citeauthoryear{Li et~al.}{2022}]{li2022competition}
\begin{barticle}
\bauthor{\bsnm{Li}, \binits{Y.}},
\bauthor{\bsnm{Choi}, \binits{D.}},
\bauthor{\bsnm{Chung}, \binits{J.}},
\bauthor{\bsnm{Kushman}, \binits{N.}},
\bauthor{\bsnm{Schrittwieser}, \binits{J.}},
\bauthor{\bsnm{Leblond}, \binits{R.}},
\bauthor{\bsnm{Eccles}, \binits{T.}},
\bauthor{\bsnm{Keeling}, \binits{J.}},
\bauthor{\bsnm{Gimeno}, \binits{F.}},
\bauthor{\bsnm{Dal~Lago}, \binits{A.}}, \betal:
\batitle{Competition-level code generation with alphacode}.
\bjtitle{Science}
\bvolume{378}(\bissue{6624}),
\bfpage{1092}--\blpage{1097}
(\byear{2022})
\doiurl{10.1126/science.abq1158}
\end{barticle}
\endbibitem

\bibitem[\protect\citeauthoryear{Mirowski et~al.}{2022}]{mirowski2022co}
\begin{botherref}
\oauthor{\bsnm{Mirowski}, \binits{P.}},
\oauthor{\bsnm{Mathewson}, \binits{K.W.}},
\oauthor{\bsnm{Pittman}, \binits{J.}},
\oauthor{\bsnm{Evans}, \binits{R.}}:
Co-writing screenplays and theatre scripts with language models: An evaluation by industry professionals.
arXiv preprint arXiv:2209.14958
(2022)
\end{botherref}
\endbibitem

\bibitem[\protect\citeauthoryear{Lee et~al.}{2022}]{lee2022coauthor}
\begin{bchapter}
\bauthor{\bsnm{Lee}, \binits{M.}},
\bauthor{\bsnm{Liang}, \binits{P.}},
\bauthor{\bsnm{Yang}, \binits{Q.}}:
\bctitle{Coauthor: Designing a human-ai collaborative writing dataset for exploring language model capabilities}.
In: \bbtitle{Proceedings of the 2022 CHI Conference on Human Factors in Computing Systems},
pp. \bfpage{1}--\blpage{19}
(\byear{2022}).
\burl{https://doi.org/10.1145/3491102.3502030}
\end{bchapter}
\endbibitem

\bibitem[\protect\citeauthoryear{Borji}{2023}]{borji2023categorical2}
\begin{botherref}
\oauthor{\bsnm{Borji}, \binits{A.}}:
A categorical archive of chatgpt failures.
arXiv preprint arXiv:2302.03494
(2023)
\end{botherref}
\endbibitem

\bibitem[\protect\citeauthoryear{Lin et~al.}{2022}]{lin2022teaching}
\begin{barticle}
\bauthor{\bsnm{Lin}, \binits{S.}},
\bauthor{\bsnm{Hilton}, \binits{J.}},
\bauthor{\bsnm{Evans}, \binits{O.}}:
\batitle{Teaching models to express their uncertainty in words}.
\bjtitle{arXiv preprint arXiv:2205.14334}
(\byear{2022})
\doiurl{10.48550/ARXIV.2205.14334}
\end{barticle}
\endbibitem

\bibitem[\protect\citeauthoryear{Zhou et~al.}{2023}]{zhou2023navigating}
\begin{barticle}
\bauthor{\bsnm{Zhou}, \binits{K.}},
\bauthor{\bsnm{Jurafsky}, \binits{D.}},
\bauthor{\bsnm{Hashimoto}, \binits{T.}}:
\batitle{Navigating the grey area: Expressions of overconfidence and uncertainty in language models}.
\bjtitle{arXiv preprint arXiv:2302.13439}
(\byear{2023})
\doiurl{10.48550/ARXIV.2302.13439}
\end{barticle}
\endbibitem

\bibitem[\protect\citeauthoryear{{Nature Machine Intelligence}}{2023}]{NatMachIntell2023AIwriting}
\begin{barticle}
\bauthor{\bsnm{{Nature Machine Intelligence}}}:
\batitle{The ai writing on the wall}.
\bjtitle{Nature Machine Intelligence}
\bvolume{5}(\bissue{1}),
\bfpage{1}--\blpage{1}
(\byear{2023})
\doiurl{10.1038/s42256-023-00613-9}
\end{barticle}
\endbibitem

\bibitem[\protect\citeauthoryear{Han et~al.}{2022}]{han2022folio}
\begin{barticle}
\bauthor{\bsnm{Han}, \binits{S.}},
\bauthor{\bsnm{Schoelkopf}, \binits{H.}},
\bauthor{\bsnm{Zhao}, \binits{Y.}},
\bauthor{\bsnm{Qi}, \binits{Z.}},
\bauthor{\bsnm{Riddell}, \binits{M.}},
\bauthor{\bsnm{Benson}, \binits{L.}},
\bauthor{\bsnm{Sun}, \binits{L.}},
\bauthor{\bsnm{Zubova}, \binits{E.}},
\bauthor{\bsnm{Qiao}, \binits{Y.}},
\bauthor{\bsnm{Burtell}, \binits{M.}}, \betal:
\batitle{Folio: Natural language reasoning with first-order logic}.
\bjtitle{arXiv preprint arXiv:2209.00840}
(\byear{2022})
\doiurl{10.48550/ARXIV.2209.00840}
\end{barticle}
\endbibitem

\bibitem[\protect\citeauthoryear{Jakesch et~al.}{2023}]{jakesch2023co}
\begin{botherref}
\oauthor{\bsnm{Jakesch}, \binits{M.}},
\oauthor{\bsnm{Bhat}, \binits{A.}},
\oauthor{\bsnm{Buschek}, \binits{D.}},
\oauthor{\bsnm{Zalmanson}, \binits{L.}},
\oauthor{\bsnm{Naaman}, \binits{M.}}:
Co-writing with opinionated language models affects users' views.
arXiv preprint arXiv:2302.00560
(2023)
\end{botherref}
\endbibitem

\bibitem[\protect\citeauthoryear{Lin et~al.}{2021}]{lin2021truthfulqa}
\begin{barticle}
\bauthor{\bsnm{Lin}, \binits{S.}},
\bauthor{\bsnm{Hilton}, \binits{J.}},
\bauthor{\bsnm{Evans}, \binits{O.}}:
\batitle{Truthfulqa: Measuring how models mimic human falsehoods}.
\bjtitle{arXiv preprint arXiv:2109.07958}
(\byear{2021})
\doiurl{10.48550/ARXIV.2109.07958}
\end{barticle}
\endbibitem

\bibitem[\protect\citeauthoryear{Weidinger et~al.}{2021}]{weidinger2021ethical}
\begin{barticle}
\bauthor{\bsnm{Weidinger}, \binits{L.}},
\bauthor{\bsnm{Mellor}, \binits{J.}},
\bauthor{\bsnm{Rauh}, \binits{M.}},
\bauthor{\bsnm{Griffin}, \binits{C.}},
\bauthor{\bsnm{Uesato}, \binits{J.}},
\bauthor{\bsnm{Huang}, \binits{P.-S.}},
\bauthor{\bsnm{Cheng}, \binits{M.}},
\bauthor{\bsnm{Glaese}, \binits{M.}},
\bauthor{\bsnm{Balle}, \binits{B.}},
\bauthor{\bsnm{Kasirzadeh}, \binits{A.}}, \betal:
\batitle{Ethical and social risks of harm from language models}.
\bjtitle{arXiv preprint arXiv:2112.04359}
(\byear{2021})
\doiurl{10.48550/ARXIV.2112.04359}
\end{barticle}
\endbibitem

\bibitem[\protect\citeauthoryear{Zhuo et~al.}{2023}]{zhuo2023exploring}
\begin{barticle}
\bauthor{\bsnm{Zhuo}, \binits{T.Y.}},
\bauthor{\bsnm{Huang}, \binits{Y.}},
\bauthor{\bsnm{Chen}, \binits{C.}},
\bauthor{\bsnm{Xing}, \binits{Z.}}:
\batitle{Exploring ai ethics of chatgpt: A diagnostic analysis}.
\bjtitle{arXiv preprint arXiv:2301.12867}
(\byear{2023})
\doiurl{10.48550/ARXIV.2301.12867}
\end{barticle}
\endbibitem

\bibitem[\protect\citeauthoryear{Abid et~al.}{2021}]{abid2021large}
\begin{barticle}
\bauthor{\bsnm{Abid}, \binits{A.}},
\bauthor{\bsnm{Farooqi}, \binits{M.}},
\bauthor{\bsnm{Zou}, \binits{J.}}:
\batitle{Large language models associate muslims with violence}.
\bjtitle{Nature Machine Intelligence}
\bvolume{3}(\bissue{6}),
\bfpage{461}--\blpage{463}
(\byear{2021})
\end{barticle}
\endbibitem

\bibitem[\protect\citeauthoryear{Buchanan et~al.}{2021}]{buchanan2021truth}
\begin{bbook}
\bauthor{\bsnm{Buchanan}, \binits{B.}},
\bauthor{\bsnm{Lohn}, \binits{A.}},
\bauthor{\bsnm{Musser}, \binits{M.}}:
\bbtitle{Truth, Lies, and Automation: How Language Models Could Change Disinformation}.
\bpublisher{Center for Security and Emerging Technology}, \blocation{???}
(\byear{2021})
\end{bbook}
\endbibitem

\bibitem[\protect\citeauthoryear{O'Connor and ChatGPT}{2022}]{o2022open}
\begin{barticle}
\bauthor{\bsnm{O'Connor}, \binits{S.}},
\bauthor{\bsnm{ChatGPT}}:
\batitle{Open artificial intelligence platforms in nursing education: Tools for academic progress or abuse?}
\bjtitle{Nurse Education in Practice}
\bvolume{66},
\bfpage{103537}--\blpage{103537}
(\byear{2022})
\doiurl{10.1016/j.nepr.2022.103537}
\end{barticle}
\endbibitem

\bibitem[\protect\citeauthoryear{Transformer and Zhavoronkov}{2022}]{transformer2022rapamycin}
\begin{barticle}
\bauthor{\bsnm{Transformer}, \binits{C.G.P.-t.}},
\bauthor{\bsnm{Zhavoronkov}, \binits{A.}}:
\batitle{Rapamycin in the context of pascal’s wager: generative pre-trained transformer perspective}.
\bjtitle{Oncoscience}
\bvolume{9},
\bfpage{82}
(\byear{2022})
\doiurl{10.18632/oncoscience.571}
\end{barticle}
\endbibitem

\bibitem[\protect\citeauthoryear{Kung et~al.}{2022}]{Kung2022.12.19.22283643}
\begin{barticle}
\bauthor{\bsnm{Kung}, \binits{T.H.}},
\bauthor{\bsnm{Cheatham}, \binits{M.}},
\bauthor{\bsnm{ChatGPT}},
\bauthor{\bsnm{Medenilla}, \binits{A.}},
\bauthor{\bsnm{Sillos}, \binits{C.}},
\bauthor{\bsnm{De~Leon}, \binits{L.}},
\bauthor{\bsnm{Elepa{\~n}o}, \binits{C.}},
\bauthor{\bsnm{Madriaga}, \binits{M.}},
\bauthor{\bsnm{Aggabao}, \binits{R.}},
\bauthor{\bsnm{Diaz-Candido}, \binits{G.}},
\bauthor{\bsnm{Maningo}, \binits{J.}},
\bauthor{\bsnm{Tseng}, \binits{V.}}:
\batitle{Performance of chatgpt on usmle: Potential for ai-assisted medical education using large language models}.
\bjtitle{medRxiv}
(\byear{2022})
\doiurl{10.1101/2022.12.19.22283643}
{\href{https://arxiv.org/abs/https://www.medrxiv.org/content/early/2022/12/21/2022.12.19.22283643.full.pdf}{{https://www.medrxiv.org/content/early/2022/12/21/2022.12.19.22283643.full.pdf}}}
\end{barticle}
\endbibitem

\bibitem[\protect\citeauthoryear{Anon}{2023}]{anon2023tools}
\begin{barticle}
\bauthor{\bsnm{Anon}}:
\batitle{Tools such as chatgpt threaten transparent science; here are our ground rules for their use}.
\bjtitle{Nature}
\bvolume{613},
\bfpage{612}
(\byear{2023})
\end{barticle}
\endbibitem

\bibitem[\protect\citeauthoryear{Thorp}{2023}]{thorp2023chatgpt}
\begin{botherref}
\oauthor{\bsnm{Thorp}, \binits{H.H.}}:
ChatGPT is fun, but not an author.
American Association for the Advancement of Science
(2023)
\end{botherref}
\endbibitem

\bibitem[\protect\citeauthoryear{Frewer et~al.}{1998}]{frewer1998understanding}
\begin{barticle}
\bauthor{\bsnm{Frewer}, \binits{L.J.}},
\bauthor{\bsnm{Howard}, \binits{C.}},
\bauthor{\bsnm{Shepherd}, \binits{R.}}:
\batitle{Understanding public attitudes to technology}.
\bjtitle{Journal of Risk Research}
\bvolume{1}(\bissue{3}),
\bfpage{221}--\blpage{235}
(\byear{1998})
\end{barticle}
\endbibitem

\bibitem[\protect\citeauthoryear{de~Cosmo et~al.}{2021}]{de2021role}
\begin{barticle}
\bauthor{\bsnm{Cosmo}, \binits{L.M.}},
\bauthor{\bsnm{Piper}, \binits{L.}},
\bauthor{\bsnm{Di~Vittorio}, \binits{A.}}:
\batitle{The role of attitude toward chatbots and privacy concern on the relationship between attitude toward mobile advertising and behavioral intent to use chatbots}.
\bjtitle{Italian Journal of Marketing}
\bvolume{2021},
\bfpage{83}--\blpage{102}
(\byear{2021})
\end{barticle}
\endbibitem

\bibitem[\protect\citeauthoryear{Bii et~al.}{2018}]{bii2018teacher}
\begin{barticle}
\bauthor{\bsnm{Bii}, \binits{P.}},
\bauthor{\bsnm{Too}, \binits{J.}},
\bauthor{\bsnm{Mukwa}, \binits{C.}}:
\batitle{Teacher attitude towards use of chatbots in routine teaching.}
\bjtitle{Universal Journal of Educational Research}
\bvolume{6}(\bissue{7}),
\bfpage{1586}--\blpage{1597}
(\byear{2018})
\end{barticle}
\endbibitem

\bibitem[\protect\citeauthoryear{Abdullah et~al.}{2022}]{abdullah2022chatgpt}
\begin{bchapter}
\bauthor{\bsnm{Abdullah}, \binits{M.}},
\bauthor{\bsnm{Madain}, \binits{A.}},
\bauthor{\bsnm{Jararweh}, \binits{Y.}}:
\bctitle{Chatgpt: Fundamentals, applications and social impacts}.
In: \bbtitle{2022 Ninth International Conference on Social Networks Analysis, Management and Security (SNAMS)},
pp. \bfpage{1}--\blpage{8}
(\byear{2022}).
\bcomment{IEEE}
\end{bchapter}
\endbibitem

\bibitem[\protect\citeauthoryear{Dwivedi et~al.}{2023}]{dwivedi2023so}
\begin{barticle}
\bauthor{\bsnm{Dwivedi}, \binits{Y.K.}},
\bauthor{\bsnm{Kshetri}, \binits{N.}},
\bauthor{\bsnm{Hughes}, \binits{L.}},
\bauthor{\bsnm{Slade}, \binits{E.L.}},
\bauthor{\bsnm{Jeyaraj}, \binits{A.}},
\bauthor{\bsnm{Kar}, \binits{A.K.}},
\bauthor{\bsnm{Baabdullah}, \binits{A.M.}},
\bauthor{\bsnm{Koohang}, \binits{A.}},
\bauthor{\bsnm{Raghavan}, \binits{V.}},
\bauthor{\bsnm{Ahuja}, \binits{M.}}, \betal:
\batitle{“so what if chatgpt wrote it?” multidisciplinary perspectives on opportunities, challenges and implications of generative conversational ai for research, practice and policy}.
\bjtitle{International Journal of Information Management}
\bvolume{71},
\bfpage{102642}
(\byear{2023})
\end{barticle}
\endbibitem

\bibitem[\protect\citeauthoryear{Biswas}{2023}]{biswas2023role}
\begin{botherref}
\oauthor{\bsnm{Biswas}, \binits{S.S.}}:
Role of chat gpt in public health.
Annals of Biomedical Engineering,
1--2
(2023)
\end{botherref}
\endbibitem

\bibitem[\protect\citeauthoryear{Tlili et~al.}{2023}]{tlili2023if}
\begin{barticle}
\bauthor{\bsnm{Tlili}, \binits{A.}},
\bauthor{\bsnm{Shehata}, \binits{B.}},
\bauthor{\bsnm{Adarkwah}, \binits{M.A.}},
\bauthor{\bsnm{Bozkurt}, \binits{A.}},
\bauthor{\bsnm{Hickey}, \binits{D.T.}},
\bauthor{\bsnm{Huang}, \binits{R.}},
\bauthor{\bsnm{Agyemang}, \binits{B.}}:
\batitle{What if the devil is my guardian angel: Chatgpt as a case study of using chatbots in education}.
\bjtitle{Smart Learning Environments}
\bvolume{10}(\bissue{1}),
\bfpage{15}
(\byear{2023})
\end{barticle}
\endbibitem

\bibitem[\protect\citeauthoryear{Dempere et~al.}{2023}]{dempere2023impact}
\begin{barticle}
\bauthor{\bsnm{Dempere}, \binits{J.}},
\bauthor{\bsnm{Modugu}, \binits{K.P.}},
\bauthor{\bsnm{Hesham}, \binits{A.}},
\bauthor{\bsnm{Ramasamy}, \binits{L.}}:
\batitle{The impact of chatgpt on higher education}.
\bjtitle{Dempere J, Modugu K, Hesham A and Ramasamy LK (2023) The impact of ChatGPT on higher education. Front. Educ}
\bvolume{8},
\bfpage{1206936}
(\byear{2023})
\end{barticle}
\endbibitem

\bibitem[\protect\citeauthoryear{Shoufan}{2023}]{shoufan2023exploring}
\begin{botherref}
\oauthor{\bsnm{Shoufan}, \binits{A.}}:
Exploring students’ perceptions of chatgpt: Thematic analysis and follow-up survey.
IEEE Access
(2023)
\end{botherref}
\endbibitem

\bibitem[\protect\citeauthoryear{Tounsi et~al.}{2023}]{tounsi2023exploring}
\begin{bchapter}
\bauthor{\bsnm{Tounsi}, \binits{A.}},
\bauthor{\bsnm{Elkefi}, \binits{S.}},
\bauthor{\bsnm{Bhar}, \binits{S.L.}}:
\bctitle{Exploring the reactions of early users of chatgpt to the tool using twitter data: Sentiment and topic analyses}.
In: \bbtitle{2023 IEEE International Conference on Advanced Systems and Emergent Technologies (IC\_ASET)},
pp. \bfpage{1}--\blpage{6}
(\byear{2023}).
\bcomment{IEEE}.
\burl{https://ieeexplore.ieee.org/abstract/document/10150870}
\end{bchapter}
\endbibitem

\bibitem[\protect\citeauthoryear{Korkmaz et~al.}{2023}]{korkmaz2023analyzing}
\begin{barticle}
\bauthor{\bsnm{Korkmaz}, \binits{A.}},
\bauthor{\bsnm{Akt{\"u}rk}, \binits{C.}},
\bauthor{\bsnm{TALAN}, \binits{T.}}:
\batitle{Analyzing the user's sentiments of chatgpt using twitter data}.
\bjtitle{Iraqi Journal For Computer Science and Mathematics}
\bvolume{4}(\bissue{2}),
\bfpage{202}--\blpage{214}
(\byear{2023})
\end{barticle}
\endbibitem

\bibitem[\protect\citeauthoryear{Haque et~al.}{2022}]{haque2022think}
\begin{botherref}
\oauthor{\bsnm{Haque}, \binits{M.U.}},
\oauthor{\bsnm{Dharmadasa}, \binits{I.}},
\oauthor{\bsnm{Sworna}, \binits{Z.T.}},
\oauthor{\bsnm{Rajapakse}, \binits{R.N.}},
\oauthor{\bsnm{Ahmad}, \binits{H.}}:
" i think this is the most disruptive technology": Exploring sentiments of chatgpt early adopters using twitter data.
arXiv preprint arXiv:2212.05856
(2022)
\end{botherref}
\endbibitem

\bibitem[\protect\citeauthoryear{Taecharungroj}{2023}]{taecharungroj2023can}
\begin{barticle}
\bauthor{\bsnm{Taecharungroj}, \binits{V.}}:
\batitle{“what can chatgpt do?” analyzing early reactions to the innovative ai chatbot on twitter}.
\bjtitle{Big Data and Cognitive Computing}
\bvolume{7}(\bissue{1}),
\bfpage{35}
(\byear{2023})
\end{barticle}
\endbibitem

\bibitem[\protect\citeauthoryear{Leiter et~al.}{2023}]{leiter2023chatgpt}
\begin{botherref}
\oauthor{\bsnm{Leiter}, \binits{C.}},
\oauthor{\bsnm{Zhang}, \binits{R.}},
\oauthor{\bsnm{Chen}, \binits{Y.}},
\oauthor{\bsnm{Belouadi}, \binits{J.}},
\oauthor{\bsnm{Larionov}, \binits{D.}},
\oauthor{\bsnm{Fresen}, \binits{V.}},
\oauthor{\bsnm{Eger}, \binits{S.}}:
Chatgpt: A meta-analysis after 2.5 months.
arXiv preprint arXiv:2302.13795
(2023)
\end{botherref}
\endbibitem

\bibitem[\protect\citeauthoryear{Blei et~al.}{2003}]{blei2003latent}
\begin{barticle}
\bauthor{\bsnm{Blei}, \binits{D.M.}},
\bauthor{\bsnm{Ng}, \binits{A.Y.}},
\bauthor{\bsnm{Jordan}, \binits{M.I.}}:
\batitle{Latent dirichlet allocation}.
\bjtitle{Journal of machine Learning research}
\bvolume{3}(\bissue{Jan}),
\bfpage{993}--\blpage{1022}
(\byear{2003})
\end{barticle}
\endbibitem

\bibitem[\protect\citeauthoryear{Okey et~al.}{2023}]{okey2023investigating}
\begin{barticle}
\bauthor{\bsnm{Okey}, \binits{O.D.}},
\bauthor{\bsnm{Udo}, \binits{E.U.}},
\bauthor{\bsnm{Rosa}, \binits{R.L.}},
\bauthor{\bsnm{Rodr{\'\i}guez}, \binits{D.Z.}},
\bauthor{\bsnm{Kleinschmidt}, \binits{J.H.}}:
\batitle{Investigating chatgpt and cybersecurity: A perspective on topic modeling and sentiment analysis}.
\bjtitle{Computers \& Security}
\bvolume{135},
\bfpage{103476}
(\byear{2023})
\end{barticle}
\endbibitem

\bibitem[\protect\citeauthoryear{Praveen and Vajrobol}{2023}]{praveen2023understanding}
\begin{botherref}
\oauthor{\bsnm{Praveen}, \binits{S.}},
\oauthor{\bsnm{Vajrobol}, \binits{V.}}:
Understanding the perceptions of healthcare researchers regarding chatgpt: a study based on bidirectional encoder representation from transformers (bert) sentiment analysis and topic modeling.
Annals of Biomedical Engineering,
1--3
(2023)
\end{botherref}
\endbibitem

\bibitem[\protect\citeauthoryear{Li et~al.}{2023}]{li2023chatgpt}
\begin{botherref}
\oauthor{\bsnm{Li}, \binits{L.}},
\oauthor{\bsnm{Ma}, \binits{Z.}},
\oauthor{\bsnm{Fan}, \binits{L.}},
\oauthor{\bsnm{Lee}, \binits{S.}},
\oauthor{\bsnm{Yu}, \binits{H.}},
\oauthor{\bsnm{Hemphill}, \binits{L.}}:
Chatgpt in education: A discourse analysis of worries and concerns on social media.
arXiv preprint arXiv:2305.02201
(2023)
\end{botherref}
\endbibitem

\bibitem[\protect\citeauthoryear{F{\"u}tterer et~al.}{2023}]{futterer2023chatgpt}
\begin{botherref}
\oauthor{\bsnm{F{\"u}tterer}, \binits{T.}},
\oauthor{\bsnm{Fischer}, \binits{C.}},
\oauthor{\bsnm{Alekseeva}, \binits{A.}},
\oauthor{\bsnm{Chen}, \binits{X.}},
\oauthor{\bsnm{Tate}, \binits{T.}},
\oauthor{\bsnm{Warschauer}, \binits{M.}},
\oauthor{\bsnm{Gerjets}, \binits{P.}}:
Chatgpt in education: Global reactions to ai innovations
(2023)
\end{botherref}
\endbibitem

\bibitem[\protect\citeauthoryear{Hutto and Gilbert}{2014}]{hutto2014vader}
\begin{bchapter}
\bauthor{\bsnm{Hutto}, \binits{C.}},
\bauthor{\bsnm{Gilbert}, \binits{E.}}:
\bctitle{Vader: A parsimonious rule-based model for sentiment analysis of social media text}.
In: \bbtitle{Proceedings of the International AAAI Conference on Web and Social Media},
vol. \bseriesno{8},
pp. \bfpage{216}--\blpage{225}
(\byear{2014}).
\burl{https://doi.org/10.1609/icwsm.v8i1.14550}
\end{bchapter}
\endbibitem

\bibitem[\protect\citeauthoryear{Barbieri et~al.}{2020}]{barbieri2020tweeteval}
\begin{botherref}
\oauthor{\bsnm{Barbieri}, \binits{F.}},
\oauthor{\bsnm{Camacho-Collados}, \binits{J.}},
\oauthor{\bsnm{Neves}, \binits{L.}},
\oauthor{\bsnm{Espinosa-Anke}, \binits{L.}}:
Tweeteval: Unified benchmark and comparative evaluation for tweet classification.
arXiv preprint arXiv:2010.12421
(2020)
\end{botherref}
\endbibitem

\bibitem[\protect\citeauthoryear{Loureiro et~al.}{2022}]{loureiro2022timelms}
\begin{botherref}
\oauthor{\bsnm{Loureiro}, \binits{D.}},
\oauthor{\bsnm{Barbieri}, \binits{F.}},
\oauthor{\bsnm{Neves}, \binits{L.}},
\oauthor{\bsnm{Anke}, \binits{L.E.}},
\oauthor{\bsnm{Camacho-Collados}, \binits{J.}}:
Timelms: Diachronic language models from twitter.
arXiv preprint arXiv:2202.03829
(2022)
\end{botherref}
\endbibitem

\bibitem[\protect\citeauthoryear{Broder}{1997}]{broder1997resemblance}
\begin{bchapter}
\bauthor{\bsnm{Broder}, \binits{A.Z.}}:
\bctitle{On the resemblance and containment of documents}.
In: \bbtitle{Proceedings. Compression and Complexity of SEQUENCES 1997 (Cat. No. 97TB100171)},
pp. \bfpage{21}--\blpage{29}
(\byear{1997}).
\bcomment{IEEE}
\end{bchapter}
\endbibitem

\bibitem[\protect\citeauthoryear{Barbieri et~al.}{2022}]{barbieri2022xlm}
\begin{bchapter}
\bauthor{\bsnm{Barbieri}, \binits{F.}},
\bauthor{\bsnm{Anke}, \binits{L.E.}},
\bauthor{\bsnm{Camacho-Collados}, \binits{J.}}:
\bctitle{Xlm-t: Multilingual language models in twitter for sentiment analysis and beyond}.
In: \bbtitle{Proceedings of the Thirteenth Language Resources and Evaluation Conference},
pp. \bfpage{258}--\blpage{266}
(\byear{2022}).
\burl{https://doi.org/10.48550/arXiv.2104.12250}
\end{bchapter}
\endbibitem

\bibitem[\protect\citeauthoryear{Grootendorst}{2022}]{grootendorst2022bertopic}
\begin{botherref}
\oauthor{\bsnm{Grootendorst}, \binits{M.}}:
Bertopic: Neural topic modeling with a class-based tf-idf procedure.
arXiv preprint arXiv:2203.05794
(2022)
\end{botherref}
\endbibitem

\bibitem[\protect\citeauthoryear{Cheng and Jiang}{2022}]{cheng2022customer}
\begin{barticle}
\bauthor{\bsnm{Cheng}, \binits{Y.}},
\bauthor{\bsnm{Jiang}, \binits{H.}}:
\batitle{Customer--brand relationship in the era of artificial intelligence: understanding the role of chatbot marketing efforts}.
\bjtitle{Journal of Product \& Brand Management}
\bvolume{31}(\bissue{2}),
\bfpage{252}--\blpage{264}
(\byear{2022})
\end{barticle}
\endbibitem

\bibitem[\protect\citeauthoryear{Frith}{2023}]{frith2023chatgpt}
\begin{barticle}
\bauthor{\bsnm{Frith}, \binits{K.H.}}:
\batitle{Chatgpt: Disruptive educational technology}.
\bjtitle{Nursing Education Perspectives}
\bvolume{44}(\bissue{3}),
\bfpage{198}--\blpage{199}
(\byear{2023})
\end{barticle}
\endbibitem

\bibitem[\protect\citeauthoryear{Chow et~al.}{2023}]{chow2023impact}
\begin{botherref}
\oauthor{\bsnm{Chow}, \binits{J.C.}},
\oauthor{\bsnm{Sanders}, \binits{L.}},
\oauthor{\bsnm{Li}, \binits{K.}}:
Impact of chatgpt on medical chatbots as a disruptive technology.
Frontiers in Artificial Intelligence
\textbf{6}
(2023)
\end{botherref}
\endbibitem

\bibitem[\protect\citeauthoryear{Choi et~al.}{2023}]{choi2023chatgpt}
\begin{botherref}
\oauthor{\bsnm{Choi}, \binits{J.H.}},
\oauthor{\bsnm{Hickman}, \binits{K.E.}},
\oauthor{\bsnm{Monahan}, \binits{A.}},
\oauthor{\bsnm{Schwarcz}, \binits{D.}}:
Chatgpt goes to law school.
Available at SSRN
(2023)
\end{botherref}
\endbibitem

\bibitem[\protect\citeauthoryear{Tubishat et~al.}{2023}]{tubishat2023sentiment}
\begin{bchapter}
\bauthor{\bsnm{Tubishat}, \binits{M.}},
\bauthor{\bsnm{Al-Obeidat}, \binits{F.}},
\bauthor{\bsnm{Shuhaiber}, \binits{A.}}:
\bctitle{Sentiment analysis of using chatgpt in education}.
In: \bbtitle{2023 International Conference on Smart Applications, Communications and Networking (SmartNets)},
pp. \bfpage{1}--\blpage{7}
(\byear{2023}).
\bcomment{IEEE}.
\burl{https://ieeexplore.ieee.org/abstract/document/10215977}
\end{bchapter}
\endbibitem

\bibitem[\protect\citeauthoryear{Rosenthal et~al.}{2019}]{rosenthal2019semeval}
\begin{botherref}
\oauthor{\bsnm{Rosenthal}, \binits{S.}},
\oauthor{\bsnm{Farra}, \binits{N.}},
\oauthor{\bsnm{Nakov}, \binits{P.}}:
Semeval-2017 task 4: Sentiment analysis in twitter.
arXiv preprint arXiv:1912.00741
(2019)
\end{botherref}
\endbibitem

\end{thebibliography}

\end{document}